# Recognition of handwritten *Bangla* basic characters and digits using convex hull based feature set

Nibaran Das*, Sandip Pramanik*, Subhadip Basu*, Punam Kumar Saha+, Ram Sarkar*, Mahantapas Kundu*, Mita Nasipuri*

*Computer Sc. & Engg. Dept., Jadavpur University, Kolkata-700032, India.
+Electrical and Computer Engineering, The University of Iowa Iowa City, IA 52242,USA
Corresponding author(s): nibaran@gmail.com

## Abstract

*In dealing with the problem of recognition of handwritten character patterns of varying shapes and sizes, selection of a proper feature set is important to achieve high recognition performance. The current research aims to evaluate the performance of the convex hull based feature set, i.e. 125 features in all computed over different bays attributes of the convex hull of a pattern, for effective recognition of isolated handwritten Bangla basic characters and digits. On experimentation with a database of 10000 samples, the maximum recognition rate of 76.86% is observed for handwritten Bangla characters. For Bangla numerals the maximum success rate of 99.45%. is achieved on a database of 12000 sample. The current work validates the usefulness of a new kind of feature set for recognition of handwritten Bangla basic characters and numerals*

## 1. Introduction.

Handwritten character recognition is widely considered as a benchmark problem of Pattern Recognition and Artificial Intelligence. The task of recognition can broadly be classified into two categories, viz., recognition of machine printed text and the recognition of handwritten text. Machine printed characters are uniform in size, position and pitch for any given font. In contrast, handwritten characters are non-uniform, involving variability in the writing style of different individuals. Despite these challenges, recognition of handwritten text is a popular research area for many years. Due to large varieties of potential applications like extracting data from filled in forms, automatic postal code identification, and mail sorting systems, automatic reading of bank cheques etc, handwritten character recognition is considered as an important problem. Past work on recognition of handwritten characters has been mostly found to concentrate on Roman script [1] related to English and some European languages, and scripts related to Asian languages like Chinese [2], Korean, and Japanese. Among Indian scripts, Devnagri, Tamil, Oriya and *Bangla* have started to receive attention for OCR related research in the recent years.

*Bangla*, the second most popular language in India and also the national language of *Bangla*desh, is the fifth most popular language in the world. But despite its importance and popularity, evidences of research on OCR of handwritten *Bangla* characters, as observed in the literature, are a few in numbers.

In the light of above facts, the present work concentrates on the development of an MLP based pattern classifier for recognition of handwritten *Bangla* basic characters and numerals with a newly introduced convex-hull based feature set[3]. In dealing with the problem of recognition of character patterns of varying shapes and sizes, selection of a proper feature set is important to achieve high recognition performance. The current research aims to evaluate the performance of the convex hull based feature set for effective recognition of isolated handwritten *Bangla* basic characters and digits.

The convex hull of a point-set is the smallest convex space that contains all the points belongings to the set. For a finite 2D point-set, the convex hull may be defined as the smallest convex polygon containing all the points. In our Present work, we have used Graham scan algorithm [4] for computing the convex hull of each numeric pattern. The worst-case complexity of this algorithm for a point-set containing n points is O(n log n). Convex hulls have several useful properties which make them suitable for many recognition and representation tasks.

It is true that objects which are very different in shape may have identical convex hulls, and that would be a problem if we were to exclusively use the convex hull attributes as features. Therefore, we have extracted different topological features (like bays attributes, lakes etc) of the convex hull from the character images. To extract local information from such images, each such character pattern is further divided into four sub-images based on the centroid of its convex hull. After that, new convex hulls are constructed for each such sub-image. In our method for recognition of handwritten *Bangla* basic characters and numerals, we have used simple 125 features based on different bays and lake attributes of the convex hull using MLP based classifier. The work presented here not only targets the development of a suitable feature set for recognition of handwritten *Bangla* character and digits in the feature space but also the development of an effective classification technique for dealing with the same. These have been the primary motivation behind the current work, presented in this paper.

### 1.1 A brief overview of *Bangla* script.

From the literature, it has been noticed that recently *Bangla* script has started to receive attention for OCR related research. If we analyze the *Bangla* script we have found there are 10 digits and 50 basic characters. Though there also exists 10 modifiers and around 260 compound characters in said script but our research interest, for the time being, is limited within basic characters and numerals only. Among these basic characters, 11 are vowels and 39 are consonants. Figure 1(a-c). shows the typical handwritten digit and basic character patterns of *Bangla* script. The writing style of *Bangla* is horizontal and left to right and the script is not case-sensitive.

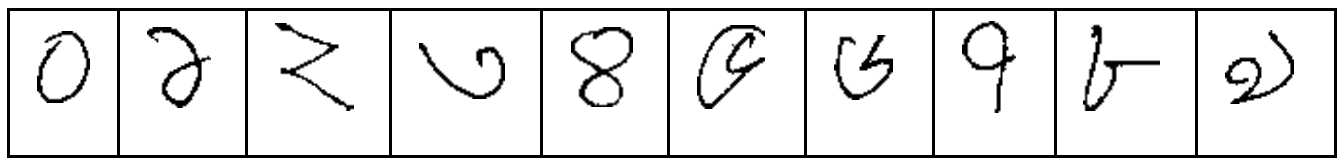

**(a)**

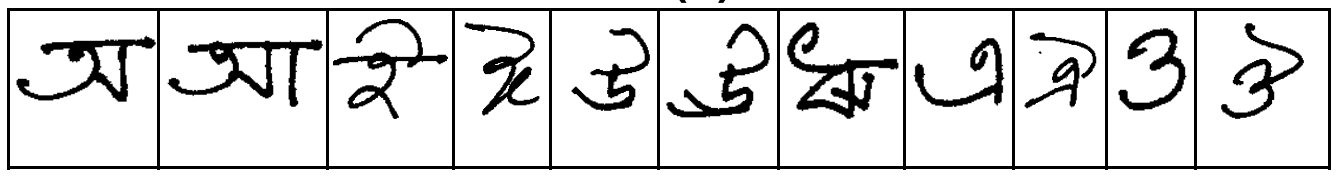

**(b)**

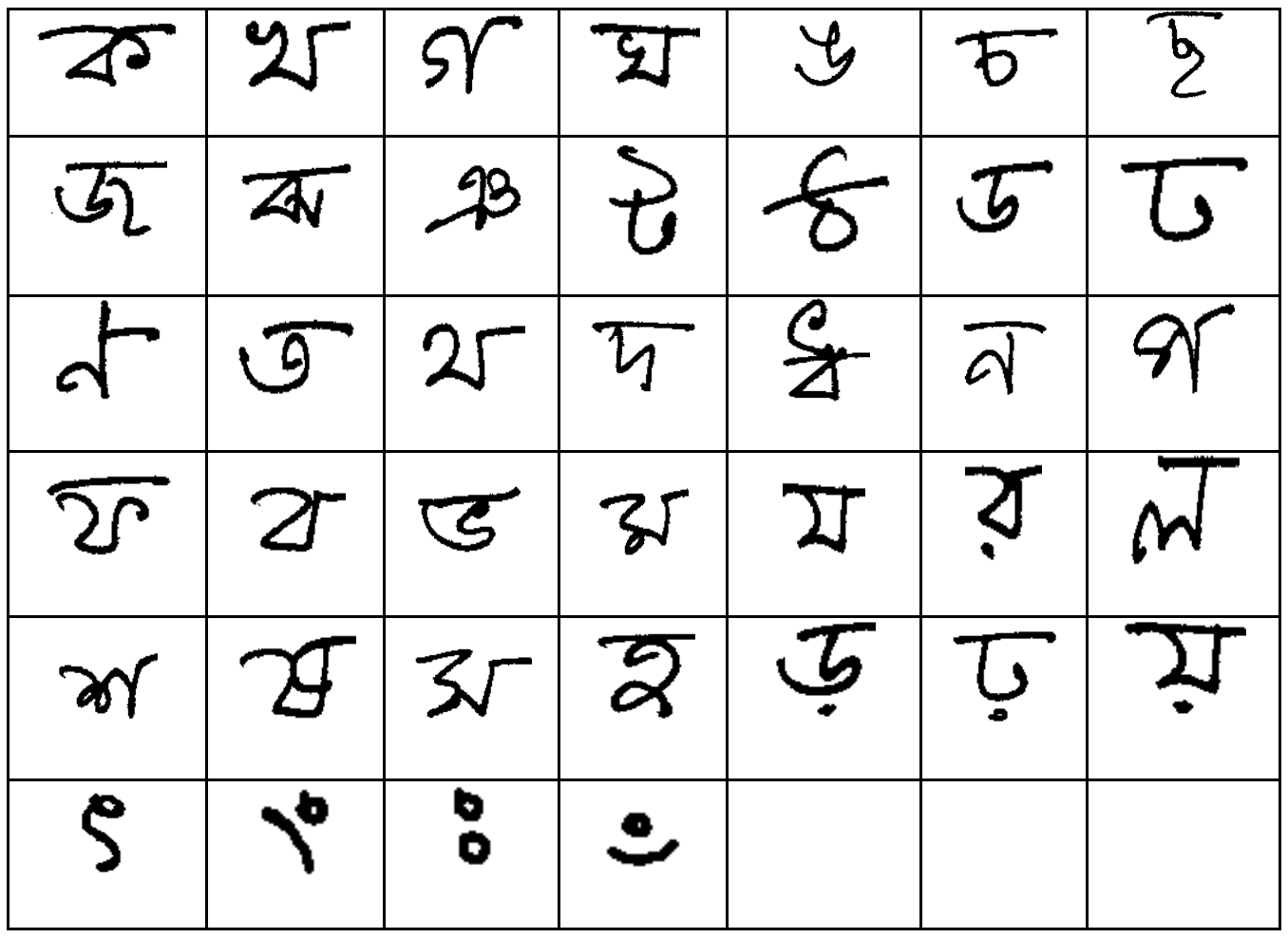

**(c)**

**Figure 1. Sample images of 50 handwritten characters of *Bangla* digit and alphabet.**
**(a). Digits of *Bangla* script**
**(b). Vowels of *Bangla* script**
**(c). Consonants of *Bangla* script**

### 1.2 Previous work.

Some of the important research contributions relating to OCR of *Bangla* characters involve a multistage approach developed by Rahman et al. [5] and an MLP classifier developed by Bhowmik et al. [6]. The major features used for the multistage approach include Matra, upper part of the character, disjoint section of the character, vertical line and double vertical line. And, for the MLP classifier, the feature set is constructed from the stroke features of characters.

In the work of S Basu et al.[7] handwritten two MLP classifiers with two different feature sets, are combined by Dempster-Shafer (DS) technique for recognize handwritten *Bangla* digit.

In [8] the authors have developed a two pass approach where in the first pass classifier performs a coarse classification, based on some global feature set. In the second pass, group classification is done on the feature set extracted from the selected local region. The regions are selected using GA with MLP classifier to refine the earlier decision by combining the local and global features for selecting the true class.

In the work of Bhattacharya et al.[9], a two-stage approach is adopted to classify 50 handwritten Basic characters and 10 numeric digits of *Bangla* script. In this approach also a coarse or a group based coarse classification of an unknown pattern in first stage is followed by a finer classification in the second stage. Based on the similarity of shapes, 57 pattern classes are identified for final classification. These pattern classes are clustered into 11 groups for coarse classification. In another work, Bhattacharya et al.,[10] proposed a similar two stage approach for recognition of 50 Basic characters of handwritten *Bangla* script. Chain-code histogram features are used in both the cases for classification through MLP based classifiers.

Oh et.al. [11] proposed a new approach for combination of multiple features in handwriting recognition problem. In the said scheme two different sets of feature vectors are designed, where one feature-set is designed to be used by all the classes and in the other set, class-dependent feature sets are designed for patterns of each class. A combination algorithm is finally developed to combine the said feature sets for classification of patterns using a neural network classifier.

S. Basu et al. [12] developed an MLP based classifier for recognition of handwritten Bangla alphabet using 76 feature set. K. Roy et al.[18] developed a quadratic classifier based approach where features are calculated from the chain code histogram for Bangla basic character recognition where features are obtained are obtained from directional chain code histogram.

References related to convex hull based features for handwritten character recognition are few in numbers. C. Gope et al. [13] presented an affine invariant point-set matching technique which measures the similarity between two point-sets by embedding them into an affine invariant feature space, utilizes the convex hull of the point-set to extract affine invariant features. R. Minhas et al. [14] extracted features in images called convex diagonal, convex quadrilateral are used for accurate image registration. Convex diagonals, convex quadrilaterals have attractive properties like easy extraction, geometric invariance and frequent occurrence. P. P. Roy et al. [15] presented a scheme towards recognition of English character in multi-scale and multi-oriented environments. Graphical document such as map consists of text lines which appear in different orientation. The feature set used here is invariant to character orientation. Circular ring and convex hull have been used along with angular information of the contour pixels of the character to make the feature rotation invariant. Circular ring and convex hull have been used to divide a character into several zones and zone wise angular histogram is computed to get higher dimensional feature for better performance. In one of our recent research contribution [3], use first successfully employed a convex hull based feature set for recognition of handwritten Roman numerals. As per best of our knowledge no convex hull based feature is used for recognition of handwritten *Bangla* basic characters and digits.

The objective of the current work is to evaluate the performances of a new convex hull based feature set on a complex handwritten character and digit data set. In that sense these newly introduced feature set is used for the first time for recognition of handwritten *Bangla* basic characters and numerals here.

## 2. Computation of convex hull.

As discussed earlier, convex hull of any binary pattern may be defined as in this work, we have used Graham scan algorithm [4] for computation of the convex hull of a binary character patterns. Apart from their computational efficiency, convex hulls are particularly suitable for affine matching as they are affine invariant. In other words, if a point-set undergoes an affine transformation, the convex hull of the point-set undergoes the same affine transformation. Also, convex hulls have local controllability, i.e. they are only locally altered by point insertions/deletions/perturbations. This is a useful property as far as noise tolerance and partial occlusions are concerned. From the green theorem [16] it can be shown that the area A of Convex hull is given by

$$A = \frac{1}{2}\sum_{i=1}^{L}(x_i y_{i+1} - x_{i+1} y_i).$$

L = Number of order vertices, (xi , yi) coordinates of the order vertices forming polygon.

Also, the centroid ( Cx , Cy ) of the convex hull can be expressed as ,

$$c_x = \frac{1}{6A}\sum_{i=1}^{L}(x_i + x_{i+1})(x_i y_{i+1} - x_{i+1} y_i)$$

$$c_y = \frac{1}{6A}\sum_{i=1}^{L}(y_i + y_{i+1})(x_i y_{i+1} - x_{i+1} y_i)$$

it is proved that the centroid ( Cx , Cy ) is affine invariant, i.e. the centroid of the affine transformed convex hull is the affine transformed centroid of the original convex hull. Convex hulls of sample patterns of Handwritten *Bangla* numerals, computed by the Graham scan algorithm, are shown in Figure .2.

The set of pixels inside the convex hull of any object pattern which does not belong to the said object is called the deficit of convexity. There may be two types of convex deficiencies viz., region(s) totally enclosed by the object, called lakes and region(s) lying between the convex hull perimeter and the object, called bays. This is illustrated in Figure 3.

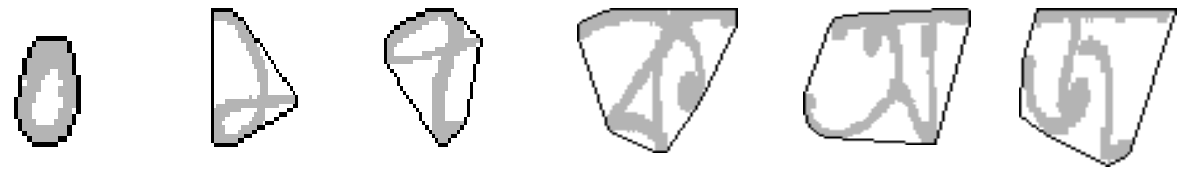

**Figure 2. Sample of some *Bangla* digit and basic characters with their corresponding convex hulls, represented in grey and black colors respectively.**

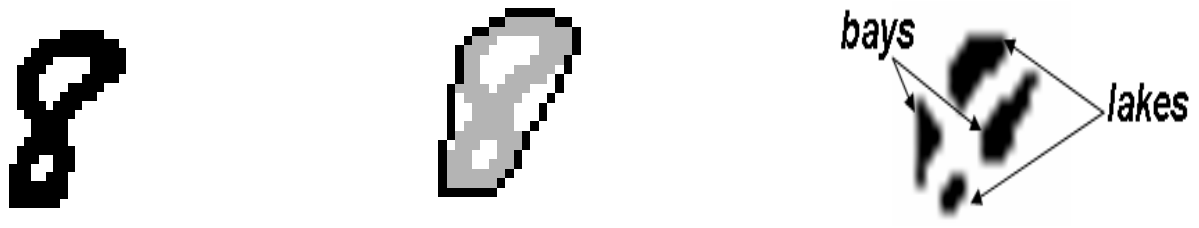

**a) Object b) Convex hull c) Bays and lake**

**Figure 3. Illustration of different convex shape descriptors for a digit image of Bangla four.**

## 3. Design of the feature set.

Any object with a non-regular shape may be represented by a collection of its topological components or features. In the current work, we have extracted several such topological features from the convex hull of handwritten *Bangla* characters.

In the current work, 25 features are designed on the basis of different bays attributes of the convex hull of handwritten *Bangla* characters. From the top, bottom, right and left boundaries of any image, column and row wise distances of data pixel from convex hull boundary are calculated as dcp. Then the maximum dcp, i.e. total no. of rows having $d_{cp} > 0$ , Average dcp, mean row co-ordinate(rx) having $d_{cp} > 0$ , total no. of rows having $d_{cp} = 0$ , number of visible bays in this direction are

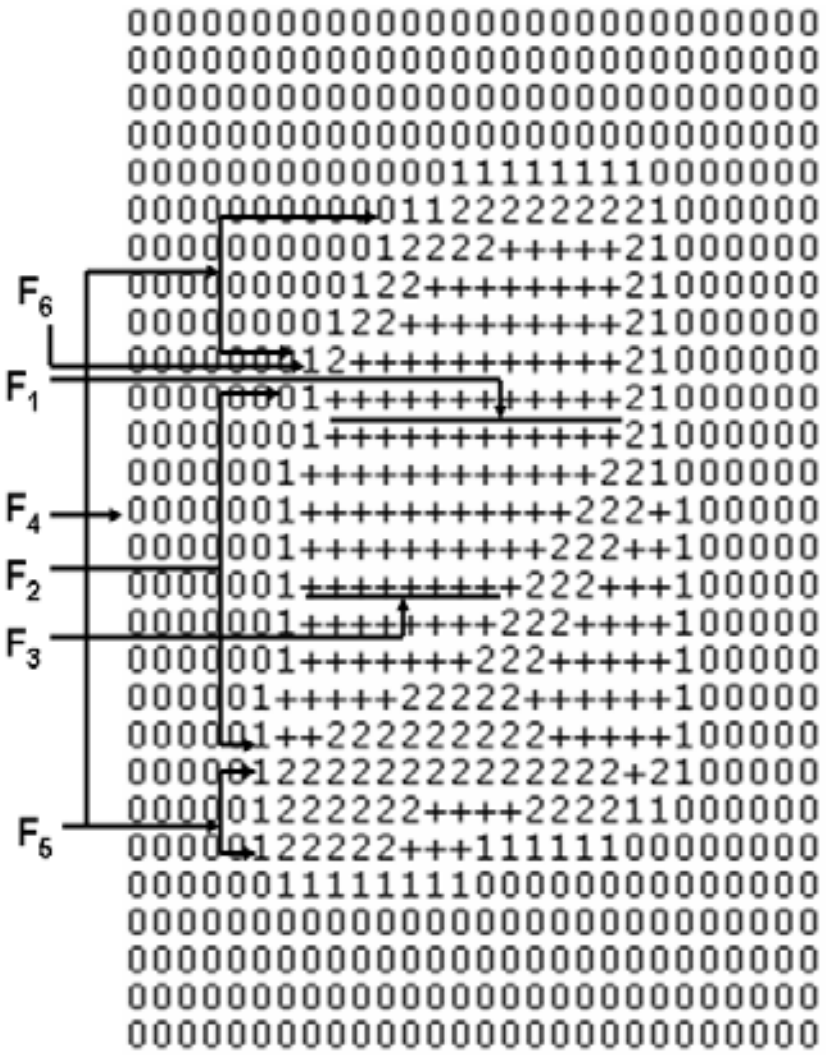

**Figure 4(a).**

| Feature No | Description | Values |
|---|---|---|
| F1 | Maximum dcp | 12 |
| F2 | Total no. of rows having dcp > 0 | 12 |
| F3 | Average dcp | 8.8 |
| F4 | Mean row co-ordinate (rx) having dcp > 0 | 14 |
| F5 | Total no. of rows having (dcp = 0) | 8 |
| F6 | Number of visible bays in this direction | 1 |
| Convex hull perimeter feature calculated as, total no. of convex hull pixels having $d_{cp} = 0$ from four sides = 41 | | |

**Table 1. Row wise from left to right**

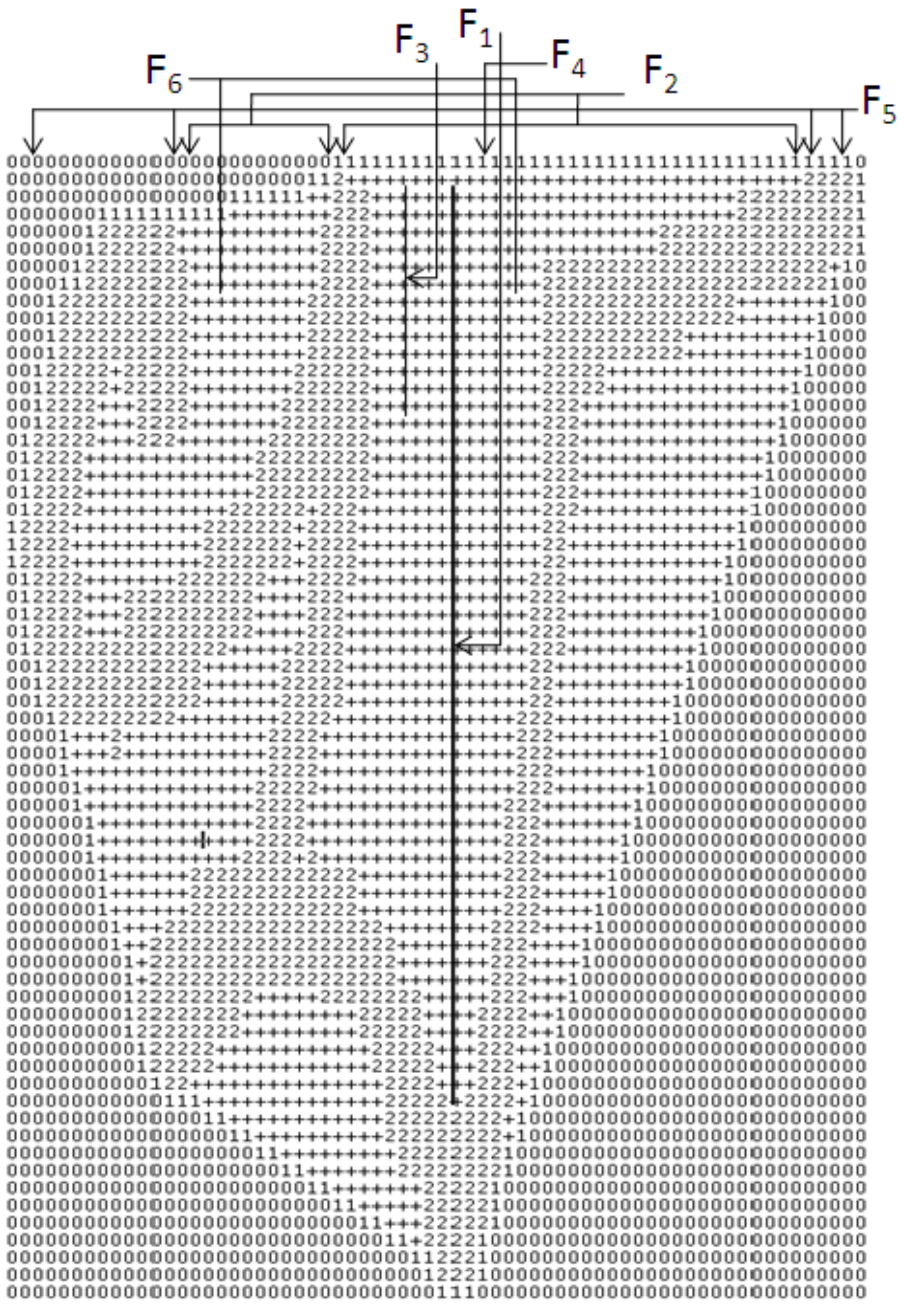

**Figure 4(b).**

**Table 2. Column wise from top to bottom**

| Feature No | Description | Values |
|---|---|---|
| F1 | Maximum dcp | 54 |
| F2 | Total no. of rows having dcp > 0 | 47 |
| F3 | Average dcp | 14.87 |
| F4 | Mean column co-ordinate (cx) having dcp > 0 | 37 |
| F5 | Total no. of rows having (dcp = 0) | 17 |
| F6 | number of visible bays in this direction | 2 |
| Convex hull perimeter feature calculated as, total no. of convex hull pixels having $d_{cp} = 0$ from four sides = 62 | | |

**Figure 4 (a). Binary image of *Bangla* Numeral "TWO"**
**(b). Binary image of *Bangla* character "KHA"**
**The label '1' represents convex hull boundary, label '2' represents image-pixels, label '+' represents bay regions within the convex hull and label '0' represents background.**

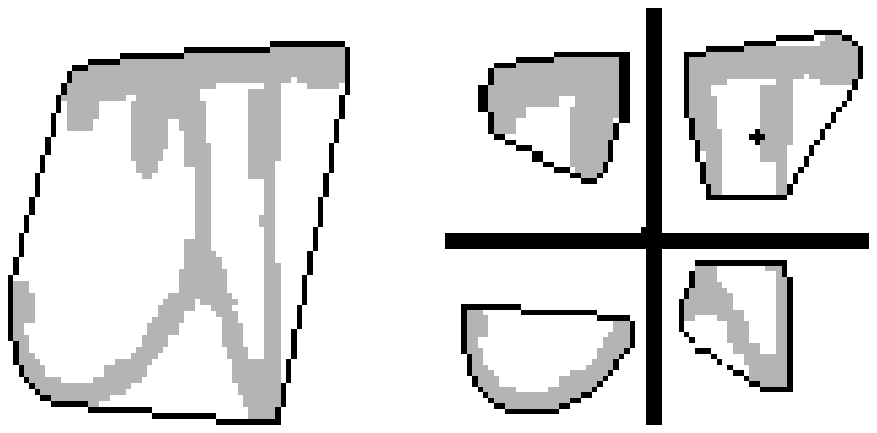

**Figure.5. Convex hulls for the sub-images of each quadrant, computed based on the centroid, are shown**

computed as six topological features. From the top, bottom, right and left boundaries of the image (6x4=24) such features are calculated. Finally, along the perimeter of the convex hull one more feature is calculated as the total number of perimeter pixels having dcp = 0. Figure 4(a-b). shows the feature extraction techniques for a *Bangla* digit and a character from the *left to right* and *top to bottom* of the image frames respectively.

As mentioned above in the Table 1-2, 25 features are extracted from the overall image based on different bays attributes of the convex hull. To extract local information, from the digit images, each such numeric pattern is further divided into four sub-images based on the centroid of its convex hull.The convex hulls are then constructed for the digit pixels within each such sub-image for computation of different topological features, as described earlier. 100 such features are computed from the 4 sub-images of each digit pattern. This make the total feature count as 125, i.e. 25 features for the overall image and 100 features in all for the four sub-images.

## 4. The MLP Classifier

In the present work, an MLP classifier [12] is employed for recognition of unknown patterns using above mentioned feature set. The MLP is a special kind of Artificial Neural Network (ANN). ANNs are developed to replicate learning and generalization abilities of human's behaviour with an attempt to model the functions of biological neural networks of the human brain.

## 5. Experimental Result:

In the present work, we have used convex hull based feature set for *Bangla* basic characters and numerals. Here, 125 features are extracted from corresponding digit and characters. For preparation of the training and the test sets of digits, a database of 12,000 samples is formed by collecting optically scanned handwritten digit samples of 10 digit symbols from each of 1200 people of different age groups and sexes. A training set of 10000 samples and a test set of 2000 samples are then formed. All the samples are converted to binary images through thresholding and scaled to 32x32 pixel images through some CG based normalization. For the alphabetic characters, a database of 10,000 samples is formed by collecting optically scanned handwritten characters specimens of 50 alphabetic symbols from each of 200 people of different age groups, sexes and professionals. A training set of 8000 samples and a test set of 2000 samples are then formed through random selection of character samples of each class from the initial database in equal numbers. All these samples are scaled to 64x64 pixels.

For the present work, Multi Layer Perceptron(MLP) with one hidden layer is chosen. This is mainly to keep the computational requirement of the same, low without affecting its function approximation capability [17]. To design the MLP for classification of handwritten Roman numerals , Back Propagation (BP) learning algorithm with learning rate ($\eta$) = 0.8 and momentum term ($\alpha$) = 0.7 is used here for training the classifier with varying number of neurons in its hidden layer. Recognition performances of the MLP for *Bangla* numerals and characters, as observed from this experiment are given in Table 3 and Table 4. Curves showing the same are also plotted in Figure 7 and Figure 8. The recognition rate of the classifier, as recorded on the test data set, initially rises as the number of neurons in its hidden layer is increased and falls after it crosses some limiting value. It reflects the fact that for some fixed training and test sets, generalization ability of an MLP improves as the number of neurons in its hidden layer is increased up to certain limiting value and any further increase in the same, thereafter, degrades this ability. It is also true for the learning ability recorded on the training set. The phenomenon is called the over-fitting problem. As observed from Table 3, the best recognition rate on test set for handwritten bangle characters is observed as 76.86% with 60 hidden neurons for handwritten *Bangla* characters. For the handwritten *Bangla* numerals the result is 99.45% with 40 hidden neurons observed in Table 4.

## 6. Conclusion.

In the current work, we have used an effective convex hull based feature set, designed for recognition of handwritten *Bangla* characters and numerals. As observed from the experiments, the current technique achieves a maximum recognition rate of 76.86% on the *Bangla* basic characters data set with no rejection. The result is comparable with some previous work [12] for the same

| Table 3 | |
|---|---|
| **No. of Hidden Neurons** | **% Success Rate** |
| 40 | 74.86 |
| 45 | 75.19 |
| 50 | 74.33 |
| 55 | 76.58 |
| **60** | **76.86** |
| 65 | 76.75 |
| 70 | 76.72 |
| 75 | 76.39 |
| 80 | 76.03 |

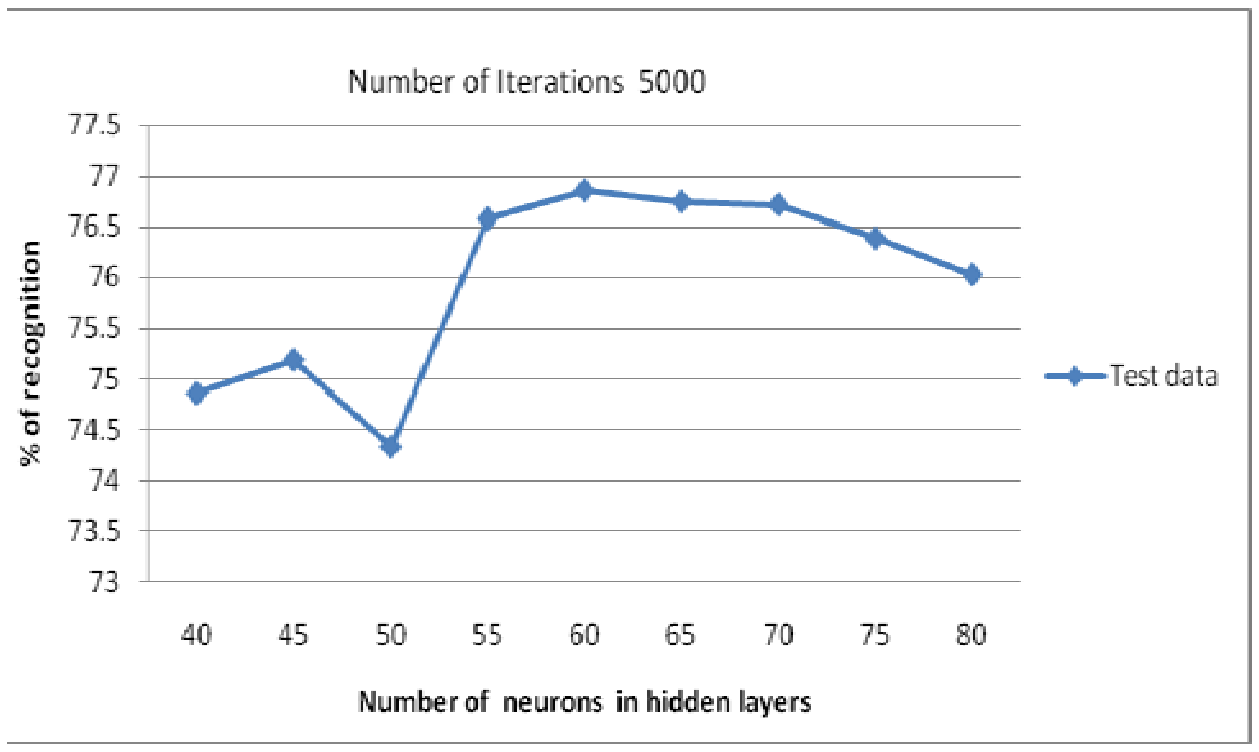

**Figure 7. Curves show variation of recognition performances of the MLP as the number of neurons in its hidden layer is increased for the test set of *Bangla* characters**

| Table 4 | |
|---|---|
| **No. of Hidden Neurons** | **% Success Rate** |
| 20 | 99.30 |
| 25 | 99.15 |
| 30 | 99.40 |
| 35 | 99.25 |
| **40** | **99.45** |
| 45 | 99.35 |
| 50 | 99.35 |
| 55 | 99.30 |
| 60 | 99.35 |
| 65 | 99.35 |

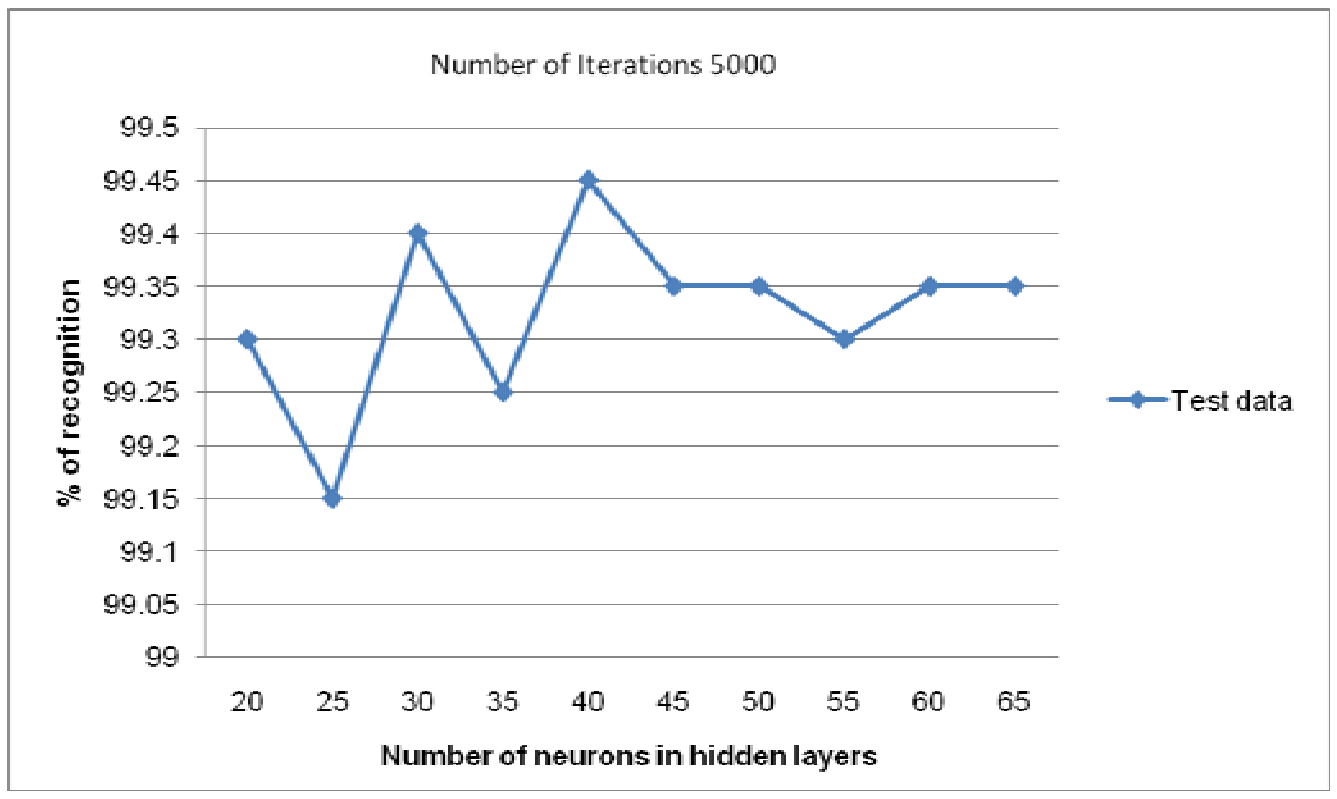

**Figure 8. Curves show variation of recognition performances of the MLP as the number of neurons in its hidden layer is increased for the test sets of Numerals**

database, available in the literature. There is no previous work published in literature using the numeral database used here. Therefore, the work cannot be directly compared with others. Still it can be said that the recognition performance, achieved here, is quite comparable with that of a contemporary work [8], mentioned before. The reliability of the system may further be enhanced by introducing rejection criteria and also by further improving the feature set. It is true that objects which are very different in shape may have identical convex hulls, and that would be a problem if we were to exclusively use the convex hull attribute as features. Therefore we have used different kind of attributed, mentioned before. We may use conventional feature set combined with these kind of features where affine invariant attributes are present makes it relevant for possible feature combination schemes for enhancement of the overall recognition performance of the pattern class.

## Acknowledgements.

Authors are thankful to the “Center for Microprocessor Application for Training Education and Research”, “Project on Storage Retrieval and Understanding of Video for Multimedia” of Computer Science & Engineering Department, Jadavpur University, for providing infrastructure facilities during progress of the work.

## References.